\documentclass[sigconf, anonymous=false]{acmart}

\usepackage{booktabs} 
\usepackage{multirow}
\usepackage{graphicx}
\usepackage{todonotes}
\usepackage{multicol}
\usepackage{makecell}

\acmArticle{4}
\acmPrice{15.00}
\begin{document}
\title{Inducing Distant Supervision in Suggestion Mining through Part-of-Speech Embeddings}
\settopmatter{printacmref=false} 
\renewcommand\footnotetextcopyrightpermission[1]{} 
\pagestyle{plain}
\author{Sapna Negi}
\affiliation{%
  \institution{Insight Centre for Data Analytics}
  \streetaddress{National University of Ireland Galway}
  \city{Galway}  
}
\email{sapna.negi@insight-centre.org}

\author{Paul Buitelaar}
\affiliation{%
  \institution{Insight Centre for Data Analytics}
  \streetaddress{National University of Ireland Galway}
  \city{Galway}  
}
\email{paul.buitelaar@insight-centre.org}

\renewcommand{\shortauthors}{B. Trovato et al.}
\begin{abstract}
Suggestion Mining refers to the task of identifying suggestion expressing sentences in a given text, which is primarily a binary sentence classification task. Suggestion mining is a young problem as compared to the other well established text classification tasks, and thus lacks large hand labeled benchmark datasets. In this work, two approaches for distant supervision are proposed, where both of them use a large silver standard dataset which we refer to as the Wiki Suggestion dataset. This dataset is constructed from sentences extracted from the articles on wikiHow and Wikipedia, serving as positive and negative instances respectively. Both the approaches use a Long Short Term Memory network based architecture to learn the classification model, but vary in their method to use the silver standard dataset. The first approach directly trains the classifier on this dataset, while the second approach only learns word level representations from this dataset. In the second approach, in addition to learning word vectors, we also learn vectors for Part of Speech (POS) tags. Interestingly, using these POS representations as input features to the neural network classifiers yields the best classification accuracy despite of their very small vocabulary as compared to the word vectors. 
\end{abstract}
\maketitle
\section{Introduction}
\begin{figure}
\centering
\includegraphics[scale=0.28]{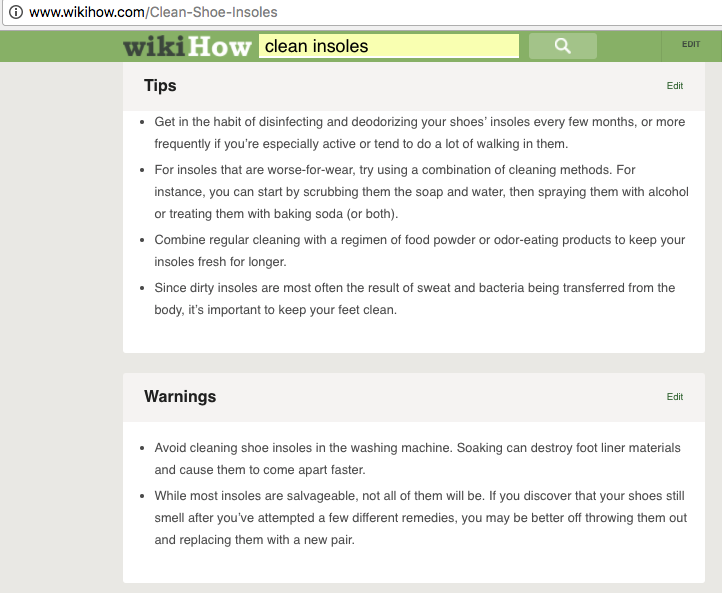}
\caption{Tips and Warning Sections of the Wikihow article on \emph{How to Clean Shoe Insoles}}
\label{wikihow}
\end{figure}

Sentences expressing advice, tips, and recommendations can often be found among the opinionated texts like reviews, blogs, tweets, discussions etc (see Table \ref{examples}). Such sentences can be collectively referred to as suggestions \cite{NegiEmnlp2015}. With the increasing availability of opinionated text, methods for automatic extraction of suggestions can be employed in different applications. The automatic extraction of suggestions from a given text is referred to as suggestion mining. \\
Although humans can infer suggestions from any given informative text, automatic inferring of suggestions might still be a far fetched task for machines. Therefore, suggestion mining has been approached with a focus on detecting the existing sentences which express suggestions in an explicit manner \cite{NegiEmnlp2015}. Studies performed in the past have defined suggestion mining as a sentence classification task, where class prediction has to be made on each sentence of a given opinionated text, classes being \emph{suggestion} and \emph{non suggestion}. State of the art opinion mining systems primarily have mostly focused on identifying sentiment polarity of the text. Suggestion mining remains a very less explored area as compared to Sentiment Analysis, specially in the context of recent advancements in neural network based approaches for learning feature representations, which is the primary focus of this work. \\
\begin{table}
\centering
\begin{tabular}{p{2.3cm} p{4.2cm}} 
\hline
{\bf Source} & {\bf Sentence}\\\hline
Hotel reviews & Be sure to specify a room at the back of the hotel.\\
Electronics reviews & I would recommend doing the upgrade to be sure you have the best chance at trouble free operation.\\
Discussion thread & If you do book your own airfare, be sure you don't have problems if Insight has to cancel the tour or reschedule it \\
Tweets & Dear Microsoft, release a new zune with your wp7 launch on the 11th. It would be smart.\\\hline
\end{tabular}
\caption{\label{examples}Examples of suggestions from different sources of opinions}
\end{table}
%
%
In a standard use of language, suggestions are expressed with specific grammatical properties, often accompanied with keywords like \emph{advice, suggest, recommend} etc. However, informal and figurative use of language on the web results in a limited performance of the manually constructed rule based classifiers \cite{Negi2016}. 
%
We observe that suggestions can be a complex combination of semantic and syntactic features. Sometimes, the features are also dependent on the domain or source of data from where suggestions are extracted, because specific topics are highly likely to appear in suggestions from a given domain. Consider two sentences from hotel reviews, \emph{You must take a long and gloomy corridor, with no decoration, which looks like you are in a basic motel.}, and \emph{You must visit the Ice Bar, 15 euros and includes a drink}. Former is annotated as a non-suggestion and latter as a suggestion, however they posses similar grammatical properties. In a statistical approach, given the domain specific training data, a classifier can model that suggestions are more likely to talk about a bar, than a corridor. \\
Neural networks are proving to be highly effective in learning automatic representations of data which is best suited for the task at hand. The existing manually labeled training datasets available for suggestion mining do not seem to be large enough to learn such representations. The problem alleviates when domain specific datasets are not available, since topics and class distribution of suggestions vary with domains. Suggestion sentences in the hand labeled datasets developed by the previous works are very sparse, ranging from 8\% to 27\% of the total annotated sentences across different datasets \cite{Negi2016}.
%
%
Therefore, we looked for existing sources of suggestion expressing sentences, and discovered that wikiHow (see Figure \ref{wikihow}) could be one potential source. In this work, we propose and evaluate two approaches of employing a large silver standard dataset assembled using the text from wikiHow and Wikipedia, which we refer to as the \emph{Wiki Suggestion Dataset}. An open domain evaluation of the approaches is also performed, where the evaluation datasets cover domains which are different from the gold standard dataset used for training the classifier. 
The contributions of this work can be summarized as:
\begin{enumerate}
\item Development of a large silver standard training dataset using wikiHow and Wikipedia.
\item Improving the classifier performance over the baseline methods, using the proposed approaches for leveraging the silver standard dataset.  
\item Experimentation with different variations of the proposed approaches providing interesting findings. Most significant one is the improved classification performance in the experiments where the words in the datasets are replaced by their POS tags, and 50 dimensional POS embeddings with a very small vocabulary are employed.
\end{enumerate}
%
%
\section{Related Work}
\subsection{Suggestion Mining}
The previous approaches for suggestion mining include linguistic rules \cite{Brun2013,Ramanand2010}, and supervised machine learning with manually determined feature types. The main algorithms used for supervised learning were, HMM and CRF \cite{Wicaksono2013}, Factorisation Machines \cite{Dong2013}, and Support Vector Machines\cite{NegiEmnlp2015}. These statistical classifiers were mostly trained and evaluated on datasets from a single domain. These studies also provided training datasets, which were all smaller than 8000 sentences, with a highly imbalanced class distribution. Only some of these datasets are publicly available. Suggestion class was the minority class in all of these datasets, with its proportion varying from 8\% to 27\% of all the sentences in the dataset.\\ 
Recently, Negi et al. \shortcite{Negi2016} stressed upon the use of neural networks for suggestion mining, and used pre-trained word embeddings with the gold standard training datasets. They compared different classifiers, including manually formulated rules, SVM with a variety of manually defined lexical, syntactic, and sentiment features, Convolutional Neural Networks, and LSTM Networks. They used 400 dimensional pre-trained embeddings, which were trained using best performing configurations for the Word2Vec algorithm \cite{baroni-dinu-kruszewski:2014:P14-1}. They performed cross-validation on a number of datasets from different domains, and their results show that LSTM Networks perform consistently better with majority of the datasets, as compared to the other methods. They also performed cross domain training and testing and the neural network based approaches using pre-trained word embeddings as input representations. Based on these findings, we also employ a LSTM based classification architecture in this work, and use the approach from Negi et al. \cite{Negi2016} as baseline. 
\subsection{Distant Supervision in Suggestion Mining}
To the best of our knowledge, only one study related to suggestion mining has focused on using distant supervision for extracting suggestions for improvement from customer feedback \cite{Moghaddam2015}. This study first defines a set of what they refer to as \emph{lexical-Part of Speech} patterns to identify suggestion reviews in a large dataset of about 50,000 eBay App reviews on App store and Google Play. The predicted suggestions were then used as positive instances to train a SVM based classifier which used bag of words features. They performed distant supervision at the review level, rather than the sentence level. They compared the results with a SVM classifier trained on a manually labeled dataset. The classifier trained on manually labeled reviews obtained higher precision, recall and F-measure values (0.38, 0.78, 0.51 respectively) on a test dataset, compared to their distant supervision method (0.32, 0.74, and 0.46 respectively). In order to extract the exact suggestion sentence within the reviews identified in the first stage, they simply used a SVM classifier with bag of words features trained on a manually annotated sentence dataset. None of their datasets are available for open research. \\
To the best of our knowledge, this work is the first to use distant supervision for suggestion mining by means of an open domain silver standard dataset and representation learning. In addition, we evaluate our approach for domain independent training.
\subsection{Learning Continuous Representations of Words}
Pennigton et al. \shortcite{pennington2014glove} provided GloVE algorithm to train general purpose word embeddings which were shown to outperform other high performance algorithms on various benchmark tasks and datasets. The GloVE outperformed algorithms included the algorithms like \emph{skip grams} and \emph{CBOW} which are two variations of the popular \emph{word2vec model}. Therefore, the pre-trained GloVE embedidngs used in \cite{pennington2014glove} are a strong baseline to evaluate the performance of the embeddings learned using the wiki suggestion dataset.\\
Training task specific embeddings have been proven to be useful for other short text classification tasks like sentiment analysis. Tang et al. \shortcite{Tang2014} trained sentiment specific word embeddings using supervised learning on a large silver standard sentiment dataset for twitter, which was labeled by means of the emoticons present in the tweets.

\section{Datasets}
In this work, we use two kinds of datasets, existing hand labeled datasets (gold standard), as well as the silver standard \emph{wiki suggestion} dataset which is introduced in this work. Table \ref{datasets} lists the statistics of the used datasets. Since it is a sentence classification task, the train and test datasets comprise of sentences labeled with their respective classes \emph{suggestion} and \emph{non-suggestion}.
\subsection{Gold Standard Datasets}
As we stated previously, some hand labeled datasets are already available from the previous works. We use the labeled datasets provided by Negi et al. \cite{NegiEmnlp2015} \cite{Negi2016}, who have performed a detailed annotation study for suggestion mining, and have set guidelines for hand labeling of datasets. In their datasets, only those sentences where suggestions are explicitly expressed are labeled as suggestions while all other sentences are labeled as non-suggestions, including the ones from which a suggestion can be inferred. For example, `I recommend the cup cakes at the bakery next door', and, `The cup cakes from the bakery
next door were delicious', are explicit and implicit forms of a suggestion respectively. The domains which are covered by these datasets include: hotel reviews, travel discussion forums, suggestion forums for software improvement, and twitter. We use all these datasets except the twitter dataset, which we consider out of scope for the current work. \\ 
We perform both in-domain as well as cross-domain training of the classifiers in order to evaluate the models for a domain independent training scenario. We employ the hotel review dataset as the gold standard training dataset in all the experiments. For in-domain evaluation of the classifier, we need a hotel domain test dataset. Since the hotel dataset is highly imbalanced with a much smaller number of suggestions than the non-suggestions, further splitting it into train and test samples would result in even lower number of distinct suggestion instances in the training sample. Therefore, we prepare a test dataset of hotel reviews obtained from the same larger sentiment analysis dataset \cite{Wachsmuth2014} from where hotel training dataset reviews were obtained, making sure that a different set of reviews are picked. Same annotation guidelines were adapted, as provided by Negi and Buitelaar \cite{NegiEmnlp2015}. A total of 5023 sentences were annotated by two annotators, out of which 400 were labeled as suggestions by both the annotators. We use a balanced subset of this dataset by only keeping 400 randomly selected non-suggestions, and removing the rest. \\ 
For the cross domain evaluation of a classifier trained on the hotel review dataset, travel discussion and software improvement datasets are used as test datasets. In this case also, we use the balanced subsets of the original versions of these datasets by removing some negative instances. \\
Table \ref{datasets} provides an overview of all the used datasets. It has been observed from the previous works that different domains tend to have different proportions of suggestions. In order to evaluate a single model for both domain specific and domain independent training, we want to avoid any bias of domain specific class distribution in the trained model. Therefore we use balanced samples in both training and evaluation datasets. We balance out the training dataset by oversampling the minority class.
\begin{table*}[t]
\centering
\begin{tabular}{|p{2.3cm}|p{4.4cm}|p{2.7cm}|p{1.8cm}|p{1.5cm}|} \hline 
{\bf Dataset} & {\bf Source/ Domain} & {\bf Sentences}& {\bf Vocabulary} & {\bf Tokens}\\ \hline
Wiki suggestion & Wikihow & 675,851 &156,768& 10,739,320 \\\cline{2-5}
 & Wikipedia & 675,851 & 548,505 & 20,863,958 \\\hline
Hotel Train & Reviews for different hotels across different cities & \makecell{Total: 7,534 \\ Suggestions: 448} & 9,349 & 1,24,721 \\\hline
Hotel Test (new) & Reviews for different hotels across different cities & \makecell{Total: 800 \\ Suggestions: 400} & 2,624 & 15,390 \\ \hline 
Travel & Travel discussion threads & \makecell{Total: 2,618 \\ Suggestions: 1,308} & 5,724 & 48,734 \\ \hline 
Software & Uservoice feedback forum for \emph{Windows} operating system, and \emph{Feedly} mobile app & \makecell{Total: 2,854 \\ Suggestions: 1,427} & 6,717 & 66,717 \\ \hline 
\end{tabular}
\caption{\label{datasets}Details of the used datasets}
\end{table*}
\subsection{Wiki Suggestion Dataset: A Silver Standard}
Due to the lack of large training datasets and a sparse distribution of suggestion instances in the existing datasets, an obvious intuition is to look for existing sources of text on the web where suggestions are already explicitly marked.\\
One such potential source for obtaining suggestion expressing sentences is web based suggestion forums for different products and services. An example of such suggestion forum is, a dedicated forum to convey the ideas for improvements in Microsoft Office 365 which is hosted by the \emph{uservoice} platform {\footnote{\url{https://office365.uservoice.com/forums/273493-office-365-admin}}}. The uservoice platform provides customer feedback management service to brands. A first impression about this suggestion forum is that it will mostly contain positive class instances for a suggestion dataset and can be directly used as a silver standard without needing any further human annotation. However, we observe that it is not the case, since text from these posts also contain many elaborative and conversational sentences in addition to the suggestions, which comprise of more than 50\% of the sentences in the posts. Therefore such datasets would also require further human annotation, although human labeling would result in training datasets with much higher percentage of suggestions compared to many other sources. \\
\begin{figure}
\centering
\includegraphics[scale=0.32]{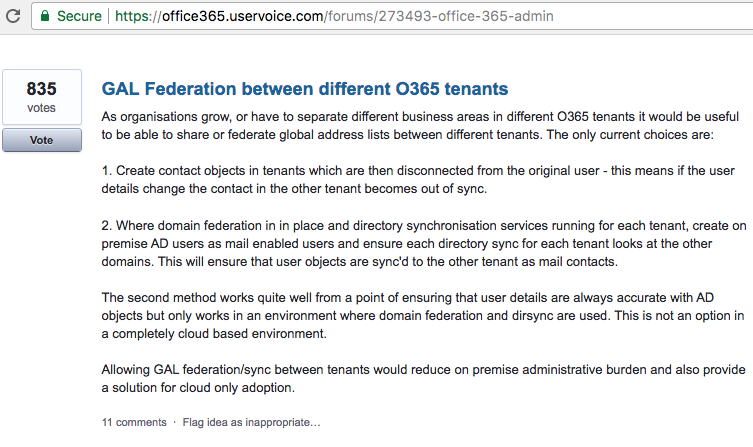}
\caption{Example of a post from the dedicated forums for users to provide suggestions for new features in brands}
\label{suggestionForum}
\end{figure}
%
WikiHow is an online wiki-style community consisting of an extensive database of how-to guides, which spans across a large variety of topics. Wikihow articles are open domain, and range over factual and non-subjective topics like \emph{Clean Shoe Insoles}, to opinionated topics like \emph{Resolve Conflict Effectively}. Each article always comprises of a \emph{Steps} section which lists the main steps of doing a certain thing (topic of the page). However, what interests us most is that there are optional sections like \emph{Tips}, \emph{Warnings}, and \emph{Community Q and A}, in addition to the steps section. The tips and warning sections contain short and to-the-point list of suggestions and advice about the topic of the article, and are very much alike the sentences which qualify for the positive instances of a suggestion mining training dataset. In order to obtain an equally large number of negative instances i.e. non-suggestions, we randomly choose equal number of sentences from Wikipedia. Wikipedia mainly contains factual descriptions, and is therefore very less likely to contain expressions of suggestions. \\
We still consider this dataset as a silver standard, due to three main reasons. Firstly, we do not perform a manual check on each sentence, and there are chances that the labels may not fully adhere to the annotation guidelines used for the gold standard datasets. Secondly, the negative sample from Wikipedia is much cleaner as compared to the ambiguous negative instances in the gold standard datasets. Thirdly, the suggestion expressions on wikihow are likely to be less varied than the gold standard datasets. \\\\
{\bf Pre-processing:} Preprocessing was only performed on the wikihow data. We extracted all the items under the tips and warnings sections. These items can sometimes be longer than one sentence. We observe that the main tip is mostly provided in the first sentence, and the following sentences contain further details and explanations, which may/ maynot carry an explicit expression of suggestion. Therefore, we perform automatic sentence splitting of each item and only retain the first sentence as a suggestion instance. The sentence splitting is performed using NLTK's \cite{Loper2002} sentence tokenizer. Before performing the sentence split, we use regular expressions to remove URLs, any further lists enclosed within an item of the main list of tips, and any content within brackets. This results in about 675,851 sentences from wikihow, and thus the final wiki suggestion dataset contains about 1.3 million sentences.  
\section{Methodology}
We propose two approaches for using the new wiki suggestion dataset for training the classifiers. As mentioned in the Related Work section, Long Short Term Memory neural networks are used for classification in both the approaches. Each approach comprises of different variations.
\subsection{Approach 1: Training data for the classifier}
In this approach, we induce distant supervision through the wiki dataset by using it as a gold standard training dataset. \\
Due to the relatively smaller size of wiki dataset and hotel training dataset, as well as our objective to perform cross domain evaluation of models, we use the full vocabulary and initial weights of pre-trained word embeddings from the GloVE\footnote{\url{https://nlp.stanford.edu/projects/glove/}} word vectors \cite{pennington2014glove}, which are trained on a very large English language corpora built by combining Wikipedia and the Gigaword corpus\footnote{\url{https://catalog.ldc.upenn.edu/ldc2011t07}} (6 billion tokens). The word vector weights from GloVE are updated each time we train the classifiers; therefore, the words which are not present in the training corpus retain the weights from GloVE. Our implementation of LSTM based classifier also saves the weights of full GloVE vocabulary when a model is saved at the end of the training process. Therefore, if the vocabulary of train and test sets do not match, the model uses the original weights from GloVE to produce the input test data representations for a trained model. We use the following dimensions of the pre-trained GloVE vectors: 200, 100, and 50.\\
This approach is evaluated with the following variations in the training data:
\begin{itemize}
\item {\bf Wiki dataset: }Only wiki suggestion dataset is used for training the model.\\
\item {\bf Wiki and hotel dataset: }Training is performed in two passes. The classifier is first pre-trained on the entire wiki dataset. The wiki trained model is then fine-tuned on the hotel train dataset.\\
\item {\bf Semantic subsample of wiki dataset: }A subsample of wiki dataset which is semantically related to the test dataset is used for training. The subset is automatically identified using the following steps:
\begin{enumerate}
\item Obtain a bag of all the unique words of the test set.
\item Obtain a vector of these bag of words by performing an addition of the corresponding word vectors (GloVe). 
\item Obtain a vector for each sentence in the wiki dataset by using the same method as above.
\item Calculate the similarity score of the test set vector with each of the wiki sentence vector.
\item Choose the top 1\% similar sentences from both the classes of the wiki dataset.
\end{enumerate}
This results in a balanced subset of about 13,500 sentences. The 100 dimensional pre-trained embeddings are used to obtain this semantic subsample. Table \ref{subsampleExamples} shows the most sentences with the highest similarity score to the respective test datasets.\\
\item {\bf Wiki subsample and hotel dataset: }Same as the variation \#2, except that the semantic subsample of wiki dataset is employed in place of the full wiki dataset.\\
\end{itemize}
{\bf Baseline: }The baseline for this approach is to use a gold standard dataset for training, which is the hotel train dataset in this work. 
\begin{table*}
\centering
\small
\begin{tabular}{|p{1.4cm}|p{4.3cm}|p{4.3cm}|p{4.3cm}|} \hline
& {\bf Hotel} & {\bf Travel}&{\bf Software}\\\hline
wikiHow& Even though most of the Universal Orlando Parks have a few areas where kids can explore and run out all their excitement, this may sometimes not be enough. & If you reacted a certain way because they did a certain thing, you'll want to bring it up tactfully. & At this time, Keek only allows users to delete their accounts from the Keek.com website, however, this feature will soon be available for Keek apps on iOS, Android, Blackberry 10, and Windows 8 devices.\\\hline
Wikipedia & Several important people of the city of Clermont had asked me to let them know when I would make the ascent. & So I thought,'ok, I'll carry on doing this for a bit and the next thing you know that's how I make my living these days". & The HD's user interface, he noted, was the first such Microsoft product to rely on text rather than icons, and it would form the basis for Windows Phone, Windows 8, Xbox and all of the company's web-based services. \\\hline
\end{tabular}
\caption{\label{subsampleExamples} Examples of sentences from the two classes in the wiki dataset with the highest similarity score with the different test datasets.}
\end{table*}
\begin{table*}
\centering
\small
\begin{tabular}{|p{2.7cm}|p{1.6cm}||p{0.6cm}|p{0.6cm}|p{0.6cm}||p{0.6cm}|p{0.6cm}|p{0.6cm}||p{0.6cm}|p{0.6cm}|p{0.6cm}|} \hline
{\bf Train data} & {\bf Embedding} & \multicolumn{3}{|c|}{\bf Hotel Test} & \multicolumn{3}{|c|}{\bf Travel} & \multicolumn{3}{|c|}{\bf Software}\\\cline{3-11}
& &P&R&F1 &P&R&F1 & P & R&F1\\\hline
\multirow{3}{*}{Hotel (Baseline)} & Glove 200 & 0.99 &0.59&0.74& 0.78 &0.26&0.40&0.70&0.28&0.40 \\
& Glove 100 & 0.95 & 0.71 & \underline{0.81} & 0.71 & 0.35 & \underline{0.47} & 0.64& 0.49 &\underline{0.55}\\
& Glove 50 & 0.97& 0.57 & 0.72 &0.72&0.28&0.40&0.69&0.39&0.50\\\hline
\multirow{3}{*}{Wiki} & Glove 200& 0.75 & 0.90& \underline{{\bf 0.82}}&0.57 & 0.70&\underline{{\bf 0.63}} &0.55&0.75 & 0.64\\ 
& Glove 100 & 0.76 &0.88&\underline{{\bf 0.82}}&0.56 & 0.69&0.62 &0.55 &0.76 & 0.64\\ 
&Glove 50 & 0.76&0.88&\underline{{\bf 0.82}}&0.56&0.70&0.62&0.56&0.79& \underline{{\bf 0.66}}\\ \hline
\multirow{3}{*}{Wiki + Hotel} & Glove 200 & 1.00 & 0.29 & 0.45 & 0.78 & 0.06 & 0.12 & 0.78 &0.07 &0.13 \\ 
& Glove 100 & 1.00 & 0.38 & \underline{0.55} & 0.83 & 0.14 & 0.23 & 0.71 & 0.21 & 0.32 \\ 
& Glove 50 & 1.00& 0.38 & \underline{0.55} &0.81& 0.18&\underline{0.29}& 0.73&0.33&\underline{0.45}\\ \hline
\multicolumn{11}{|c|}{Subsampling Wiki Dataset}\\ \hline
Wiki subsample & Glove 200 &0.25 &0.49&\underline{0.33}& 0.61 &0.46&\underline{0.53}&0.61&0.46&0.53\\  
& Glove 100 &0.99 & 0.19 &0.32& 0.70&0.14&0.23&0.52&0.74&\underline{0.61}\\
& Glove 50 &0.25 &0.49&\underline{0.33} & 0.74&0.22&0.33&0.52&0.68&0.59\\ \hline
Wiki subsample + Hotel & Glove 200&0.51&0.86&0.63&0.50&0.86&0.63&0.52&0.75&0.62\\ 
& Glove 100 & 0.99 &0.59 & \underline{0.74}&0.82&0.14&0.24&0.64&0.12&0.21\\ 
& Glove 50 & 0.50&1.00&0.67&0.50&1.00&\underline{0.67}&0.50&1.00&\underline{0.67}\\ \hline
\end{tabular}
\caption{\label{results1} Results from different dataset settings using gold standard hotel and Wiki Suggestion dataset to directly train a classifier. Precision, Recall, and F measure are provided for the positive class, i.e. the suggestion class.}
\end{table*}
%
%
\subsection{Approach 2: Training data for word embeddings}
In this approach, wiki dataset is used to induce distant supervision in the classifier model through the pre-trained word embeddings. Instead of using the standard pre-trained GloVe word vectors, we first train word embeddings using the wiki suggestion dataset, and then replace the pre-trained GloVE vectors in the previous approach with these vectors. The neural network classifier is then trained using only the gold standard dataset. Following are the different methods we employed to learn embeddings from the wiki dataset:
\begin{enumerate}
\item {\bf Wiki Suggestion Embeddings (WiSE)}: We train a classifier using the full wiki suggestion dataset, without using any pre-trained word embeddings for the initial weights. Word weights are then extracted from the embedding layer at the end of the training. The classification architecture remains same as that of the approach 1. We refer to these embeddings as the Wiki Suggestion Embeddings (WiSE). This method is different from the state of the art word co-occurrence based or neighbouring word prediction based learning (GloVE, Word2vec respectively) of embeddings, since these embeddings are learned in a supervised manner by means of a class prediction objective.\\
Specific syntactical and grammatical properties are strong features for suggestions, and suggestions across domains tend to exhibit similar syntactic properties \cite{Negi2016}. Earlier works on suggestion mining also show that POS n-gram and POS pattern based features are useful with SVM classifiers \cite{NegiEmnlp2015}. Therefore, using word vectors which strongly encode syntactic properties may be a promising direction for open domain training of suggestion mining classifiers. It has been shown earlier that the smaller dimensions of SOTA word embeddings like GloVe tend to capture syntactic properties of the words \cite{Plank2016} and perform well for syntactic analogy tasks. Table \ref{results1} also indicate that smaller dimensions of GloVe are performing better than 200 dimensions in some of the experiments. We learn embeddings for part of speech tags, which are not likely to carry any semantic information as an alternative to the standard 50 dimensional word embeddings which carry some semantic information as well.\\
%
%
%
%
%
%
\begin{table*}[h]
\centering
\small
\begin{tabular}{|p{3.2cm}|p{2.5cm}||p{0.65cm}|p{0.65cm}|p{0.65cm}||p{0.65cm}|p{0.65cm}|p{0.65cm}||p{0.65cm}|p{0.65cm}|p{0.65cm}|} \hline
{\bf Embedding} & {\bf Dimensions, Vocabulary} & \multicolumn{3}{|c|}{\bf Hotel Test} & \multicolumn{3}{|c|}{\bf Travel} & \multicolumn{3}{|c|}{\bf Software}\\\cline{3-11} 
& & P&R&F1& P&R&F1 & P & R&F1\\\hline
\multicolumn{11}{|c|}{Word Embeddings}\\ \hline
Pre-trained Glove (Baseline) & 100; 400,000 & 0.95 & 0.71& \underline{0.81}& 0.71&0.35&0.47&0.64&0.49 &\underline{0.55}\\\hline
WiSE$_w$ & 100; 339,240 & 0.26 & 0.51& 0.34& 0.25&0.50 &0.33 & 0.25& 0.50&0.33\\\hline
\multicolumn{11}{|c|}{Part of Speech Embeddings}\\ \hline
Glove (Baseline) & 50; 400,000 & 0.97&0.57&0.72&0.72&0.28&0.40&0.69&0.39 &0.50\\\hline
Glove-Wiki$_p$ & 50, 34 & 0.80 & 0.83&0.82&0.57 &0.71 & \underline{{\bf 0.63}} &0.57 &0.90 &0.70\\\hline
WiSE$_p$ & 50, 34 & 0.82 & 0.84 & \underline{0.83}& 0.57& 0.69& 0.62 & 0.58& 0.90 & \underline{\bf{0.71}} \\\hline 
\multicolumn{11}{|c|}{Word + Part of Speech Embeddings}\\ \hline
Glove + WiSE$_p$ Concat & 150; 400,000 * 34 & 0.90&0.80&{\bf 0.85}&0.62&0.57&0.59&0.61&0.81&0.70\\ \hline
\end{tabular}
\caption{\label{results2}Classification results of using different embeddings with hotel data as the training dataset. Precision, Recall, and F1 score for the suggestion class are presented.}
\end{table*}
We investigate two variants of the WiSE embeddings:
\begin{itemize}
\item WiSE$_w$: These are embeddings for words. The vocabulary comprises of words which are present in the wiki suggestion dataset, and is therefore smaller than that of the GloVE vectors. These embeddings are trained for 100 dimensions, since the baseline model also uses 100 dimensions. This variant comprises of about 339,000 words.\\
\item WiSE$_p$: These are the embeddings for Part of Speech tags. These embeddings are very light weight, with 50 dimensions and a vocabulary of size 34. These embeddings are learned by replacing all the words in the wiki corpus with their POS tags.\\
\end{itemize}
\item {\bf Glove-Wiki$_p$:} These are the POS embeddings learned using the GloVE algorithm, but trained on the wiki suggestion dataset.\\
\item {\bf Concatenation of Word and POS embeddings: }
These are the embeddings obtained by concatenating the best performing word embeddings with the best performing POS embeddings.\\
\end{enumerate}
{\bf Baseline:} The baseline for models trained using wiki dataset embeddings is the model which performed best when pre-trained GloVe vectors were used with the gold standard dataset, i.e. the model trained using 100 dimensional pre-trained GloVe (2nd model of the 1st row in Table \ref{results1}).
\subsection{Network Architecture and Hyper-parameters} 
\begin{figure}
\centering
\includegraphics[scale=0.30]{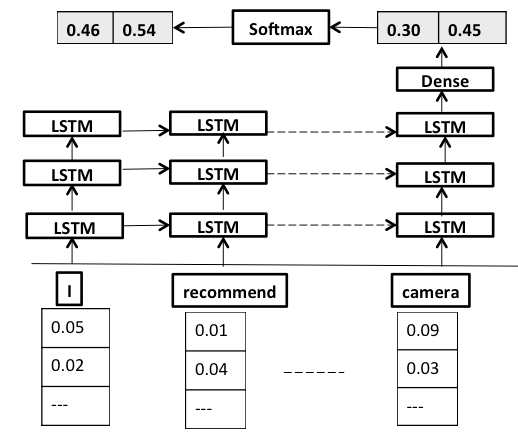}
\caption{Architecture for the LSTM based neural network classifier}
\label{architecture}
\end{figure}

We evaluated some commonly used configurations and hyper-parameter values  for short text classification. These configurations were evaluated on the baseline approach, i.e. using hotel train dataset with all three dimensions of the GloVE vectors. We chose the configurations which perform best for at-least two out of three test datasets. \\
%
The final architecture (Figure \ref{architecture}) comprises of the input layer of LSTM units, and a softmax activation is applied to the outermost dense layer of size 2. Two lstm hidden layers of size 50 and 20 units respectively. Tanh activation is used at each layer, L2 regulariser is used with the first three LSTM layers, Adam is used for optimisation, and the training was performed with 5 epochs.
Keras (theano backend), Gensim, and Scikit learn libraries are primarily used for the implementation. 
\section{Results and Analysis}
Tables \ref{results1}, \ref{results2} show results for approaches 1 and 2 respectively. Precision, Recall, and F1 score of the suggestion class (positive class) on different test datasets are reported. We compare the performance of different models on the basis of F1 score. \\
For approach 1, using the full wiki dataset as the training dataset, which is an open domain training dataset, outperforms all other variants of training datasets for all three test datasets. However, the performance improvement over baseline for the domain whose training dataset is available (hotel) is small, as compared to the other domains where domain specific training was not performed. \\ 
For approach 2, results demonstrate that the use of pre-trained POS embeddings yields better results as compared to the baseline of using pre-trained GloVe embeddings. Based on the current experiments, approach 2 appears to be a better method to employ wiki suggestion dataset for suggestion mining. The results shown in Table \ref{results2} are very interesting considering the fact that the training vocabulary for approach 2 (only POS tags) is much smaller than the standard word embeddings. However, the performances of POS embeddings learned using WiSE and GloVe methods are very close. 
%
\subsection{Qualitative and Quantitative Analysis of Wiki Dataset} 
\begin{figure}
\centering
	\includegraphics[scale=0.35]{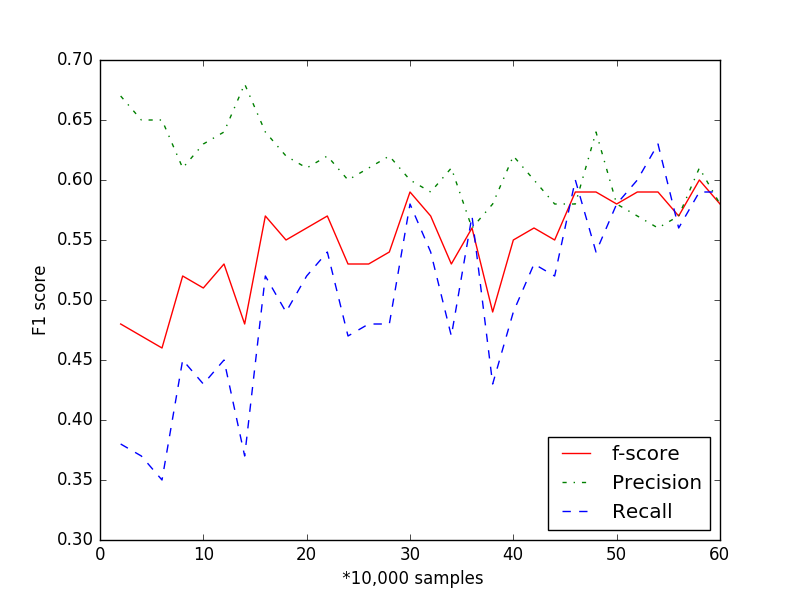}
	\caption{Training sample size of wiki dataset vs. precision, recall and F1 score on 	the travel dataset}
	\label{wikiDatasetSubsampleAdvice}
    \end{figure}

\begin{figure}
\centering
	\includegraphics[scale=0.35]{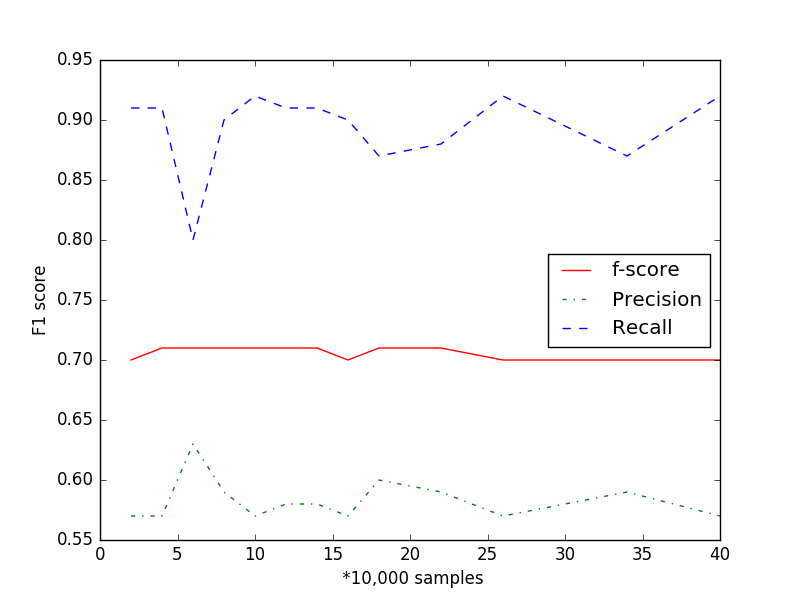}
	\caption{Training sample size for WiSE vs. precision, recall and F1 score on the 	software dataset}
	\label{wikiPOSSubsample}
\end{figure}

{\bf Approach 1:} Using full wiki dataset yielded best results, while using its subsamples significantly reduced the performance. Travel test data showed the largest improvement over baseline on using wiki dataset as training dataset. Figure \ref{wikiDatasetSubsampleAdvice} shows the change in precision, recall, and F1 score with the increasing size of wiki training dataset in approach 1, which hints that the size of the wiki dataset is the reason for its better performance over the much smaller gold standard dataset, since the results show improvements with large increases in its sample size. However, using the smaller samples of wiki dataset significantly reduced the performance despite of these samples being semantically close to the test datasets. This can also be observed in Figure \ref{wikiDatasetSubsampleAdvice} which shows the steep variations in performance within samples of comparable size.\\\\
%
%
{\bf Wiki dataset for POS embeddings: }Software test data showed the largest improvement in F1 score on using WiSE$_p$ embeddings. Figure \ref{wikiPOSSubsample} shows the change in precision, recall, and F1 score of the software test dataset when the POS embedding are learned on the different sized subsamples of wiki dataset in approach 2, which shows that the large size of dataset doesn't contribute much in this case. This is in line with our observation that the style of suggestion expressions do not vary much in this dataset. Table \ref{results4} show the results when we use only the POS version of wiki dataset to train the classifier in approach 2, but the results are much lower compared to using the gold standard (Table \ref{results2}). This hints that the model which uses POS embeddings can model syntactic variations in suggestions with a much smaller training dataset as compared to using word embeddings.\\\\
\begin{table}
\begin{tabular}
{|p{2cm}|p{0.6cm}|p{0.6cm}|p{0.6cm}|} \hline
{\bf Test Data} & {\bf P} & {\bf R} & {\bf F}\\\hline 
Hotel & 0.58 &0.50 & 0.53\\\hline
Travel & 0.52&0.55&0.53\\\hline
Software& 0.53&0.59&0.56\\\hline
\end{tabular}
\caption{\label{results4}Classification results when POS version of wiki dataset is directly used to learn the classification model, which also includes learning WiSE$_p$.}
\end{table}
\begin{table}
\begin{tabular}
{|p{1.5cm}|p{1.3cm}|p{1.5cm}|p{1.5cm}|} \hline
\multirow{2}{*}{\bf Dataset} & \multirow{2}{*}{\bf Vocab } & \multicolumn{2}{|c|}{\bf Overlapping Vocabulary} \\\cline{3-4}
& &{\bf wikiHow} &{\bf Wikipedia} \\\hline 
Hotel&2624&289&219 \\\hline
Travel&5724&838&620\\\hline
Software&6717&1643&1557\\\hline
\end{tabular}
\caption{\label{vocabulary}Overlapping vocabulary of test datasets with the wiki dataset.}
\end{table}
{\bf Wiki dataset for word embeddings: }In approach 2, WiSE$_w$ performed poorly as compared to all other embeddings. A possible reason could be that a large portion of test dataset vocabulary is not present in the wiki dataset (Table \ref{vocabulary}), therefore the vocabulary of the WiSE$_w$ embeddings is insufficient for learning task specific word embeddings.
\subsection{Word vs POS Representations}
\begin{figure}
\centering
\includegraphics[scale=0.30]{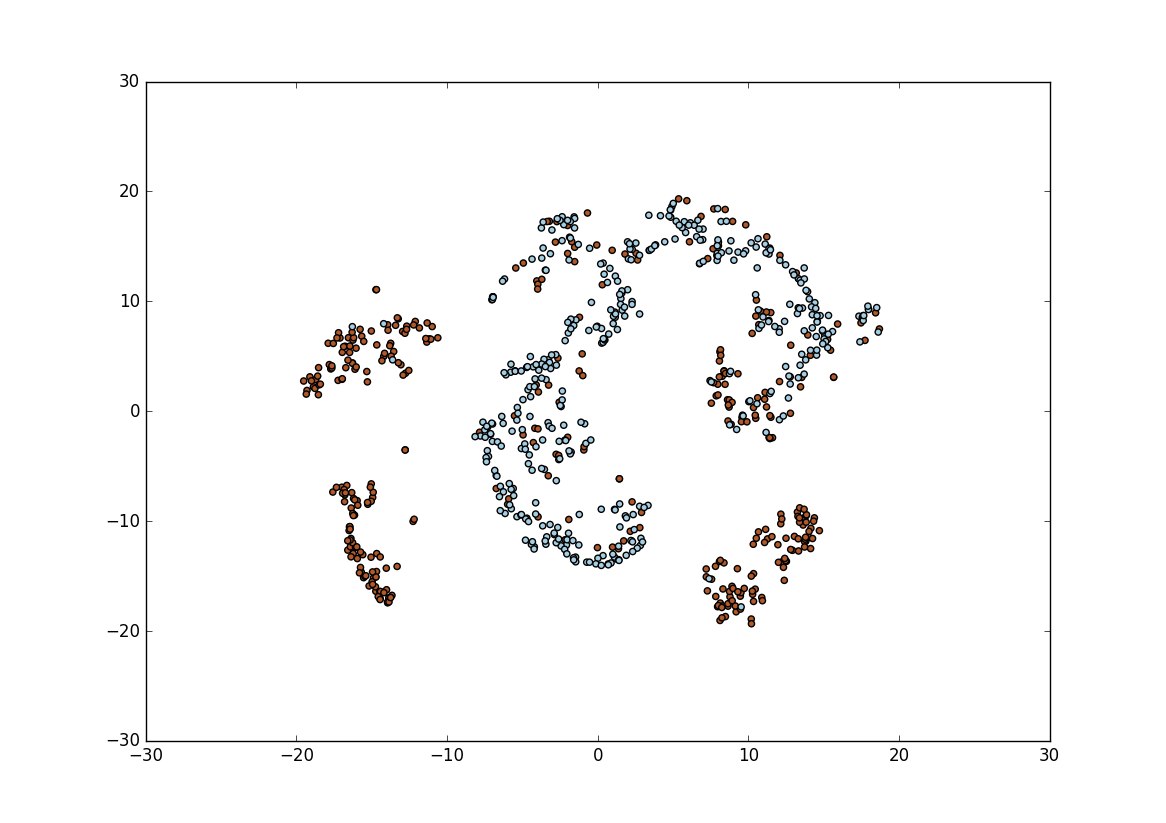}
	\caption{Sentence representations of hotel test dataset obtained from the last layer of the model trained using 100 dimensional glove embeddings and hotel train dataset.}
	\label{wordScatterPlot}
\end{figure}

\begin{figure}
\centering
\includegraphics[scale=0.30]{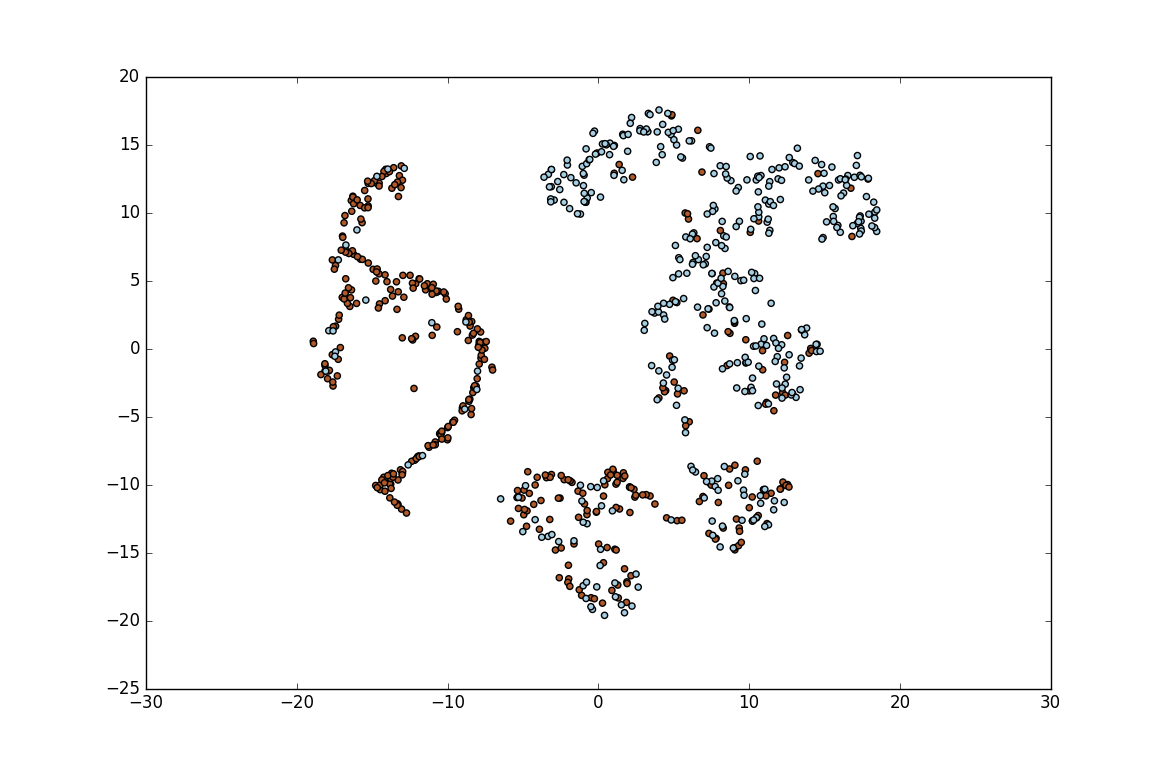}
	\caption{Sentence representations of hotel test dataset obtained from the last layer of the model which was trained using WiSE$_p$ embeddings and hotel train dataset.}
	\label{posScatterPlot}
\end{figure}

Figures \ref{wordScatterPlot} and \ref{posScatterPlot} plot sentence vectors of hotel test instances on a 2-dimensional plot using the t-sne method of dimensionality reduction \cite{Van2014}. The sentence vectors are obtained from the last LSTM layer of the models trained on the hotel train dataset, using pre-trained GloVe vs. WiSE$_p$ embeddings. Red and blue dots show the original labels for positive and negative instances respectively. \\
We can observe high variability within the representations of samples of the positive class (red) as shown by the number of clusters of positive  instances, in the GloVe embeddings based sentence representations (figure \ref{wordScatterPlot}). This may indicate the variations among the topics of suggestions. In contrast to this, WiSE$_p$ (POS embeddings) based sentence representations result in 2 major clusters for the positive instances (figure \ref{posScatterPlot}), indicating lesser syntactic variations among the suggestion instances. This may lead to an easier decision boundary in comparison to the former, and can explain the results described in the previous sub-section. 
\section{Conclusion}
Our experiments have revealed several interesting observations. Most interesting results are the high performance when using pre-trained POS embeddings with part of speech versions of the datasets, which is an extremely lightweight model. This also provides a ground for evaluating POS embeddings based approach on other sentence level tasks.\\
The two proposed approaches to use wikihow dataset perform better than the respective baselines, the POS embedding approach in particular outperforms all other variations (across both the approaches). This proves that syntax and grammar are predominant features in determining the suggestion class. However, it is also evident that in order to achieve higher classification accuracy, domain data would still be required. Last but not least, to the best of our knowledge this is the first attempt to use wikiHow for a text classification task.\\
In future we would like to investigate different neural network architectures to compose the word and POS embeddings against the simple concatenation method used in this work. 
\bibliographystyle{ACM-Reference-Format}
\bibliography{sample-bibliography} 

\end{document}